# Three Variations on Variational Autoencoders

R. I. Cukier

*Department of Chemistry, Michigan State University, East Lansing, Michigan 48824-1322, USA*
cukier@chemistry.msu.edu

Abstract

Variational autoencoders (VAEs) are one class of generative probabilistic latent-variable models designed for inference based on known data. We develop three variations on VAEs by introducing a second parameterized encoder/decoder pair and, for one variation, an additional fixed encoder. The parameters of the encoders/decoders are to be learned with a neural network. The fixed encoder is obtained by probabilistic-PCA. The variations are compared to the Evidence Lower Bound (ELBO) approximation to the original VAE. One variation leads to an Evidence Upper Bound (EUBO) that can be used in conjunction with the original ELBO to interrogate the convergence of the VAE.

## 1 Introduction

In this work, three variations on variational autoencoders (VAEs) [1-3] are suggested. VAEs along with Generative Adversarial Networks (GANs)[1, 4] form a class of generative probabilistic models that have come to the fore with the advent of deep neural networks (DNNs). They learn a probability distribution from an identically, independently distributed (i.i.d.) data collection $\mathbf{x}_i \left(i=1,2,...,N\right)$ of $N_x-$ dimensional observations. VAEs are generative; once a probability distribution $p\left(\mathbf{x}\right)$ is learned, new instances can be readily generated. An often-used strategy employs a latent, **z**-space, representation of dimension $N_z$ with $N_z \ll N_x$ embodying the idea of much redundant information in the high-dimensional $\mathbf{x} \in X$ space. For examples, the classic MNIST integer figures dataset of e.g. $N_x$=784=$28^2$ pixels can, using an autoencoder or Principle Component Analysis (PCA), be re-generated [5] with a **z**-space of dimension ~10.(1) Using a PCA- based analysis of protein conformational fluctuations where $N_x$= 856, a latent space dimension of $N_z$ ~20 already carries >90% of the total mean square fluctuation of the trajectory.[6] For proteins, the covalent bonds form a set of geometrical constraints greatly restricting the possible configuration space sampling.

(1) https://arxiv.org/pdf/1409.0473.pdf

**2 Variational autoencoders (VAEs)**

The VAE target distribution is $\log p(\mathbf{x})$. Its evaluation uses a latent space to compress information motivating the introduction of joint $p(\mathbf{x},\mathbf{z})$ and conditional $p(\mathbf{z}\,|\,\mathbf{x})$ distributions. The (unknown) latter encoder distribution provides a lower-dimensional embedding of the data. To proceed, introduce some trial conditional distribution $q(\mathbf{z}\,|\,\mathbf{x})$ to write [2]

$$\begin{aligned}
\log p(\mathbf{x}) &= \int d\mathbf{z}\, q(\mathbf{z}\,|\,\mathbf{x}) \log p(\mathbf{x}) = \int d\mathbf{z}\, q(\mathbf{z}\,|\,\mathbf{x}) \log\left[\frac{p(\mathbf{x},\mathbf{z})}{p(\mathbf{z}\,|\,\mathbf{x})}\right] \\
&= \int d\mathbf{z}\, q(\mathbf{z}\,|\,\mathbf{x}) \log\left[\frac{q(\mathbf{z}\,|\,\mathbf{x})}{p(\mathbf{z}\,|\,\mathbf{x})}\frac{p(\mathbf{x},\mathbf{z})}{q(\mathbf{z}\,|\,\mathbf{x})}\right] \\
&= \int d\mathbf{z}\, q(\mathbf{z}\,|\,\mathbf{x}) \log\left[\frac{q(\mathbf{z}\,|\,\mathbf{x})}{p(\mathbf{z}\,|\,\mathbf{x})}\right] + \int d\mathbf{z}\, q(\mathbf{z}\,|\,\mathbf{x}) \log\left[\frac{p(\mathbf{x},\mathbf{z})}{q(\mathbf{z}\,|\,\mathbf{x})}\right] \\
&= \int d\mathbf{z}\, q(\mathbf{z}\,|\,\mathbf{x}) \log\left[\frac{q(\mathbf{z}\,|\,\mathbf{x})}{p(\mathbf{z}\,|\,\mathbf{x})}\right] + \int d\mathbf{z}\, q(\mathbf{z}\,|\,\mathbf{x}) \log\left[\frac{p(\mathbf{x}\,|\,\mathbf{z})\,p(\mathbf{z})}{q(\mathbf{z}\,|\,\mathbf{x})}\right] \\
&= \int d\mathbf{z}\, q(\mathbf{z}\,|\,\mathbf{x}) \log\left[\frac{q(\mathbf{z}\,|\,\mathbf{x})}{p(\mathbf{z}\,|\,\mathbf{x})}\right] + \int d\mathbf{z}\, q(\mathbf{z}\,|\,\mathbf{x}) \log p(\mathbf{x}\,|\,\mathbf{z}) - \int d\mathbf{z}\, q(\mathbf{z}\,|\,\mathbf{x}) \log\left[\frac{q(\mathbf{z}\,|\,\mathbf{x})}{p(\mathbf{z})}\right].
\end{aligned} \tag{1}$$

In obtaining Eq. (1) we used $\int d\mathbf{z}\, q(\mathbf{z}\,|\,\mathbf{x}) = 1;\;\; \mathbf{x} \in \mathrm{X}$, appropriate to any conditional distribution. Eq. (1) can be rearranged to

$$\log p(\mathbf{x}) - D\left[q(\mathbf{z}\,|\,\mathbf{x}) \,\|\, p(\mathbf{z}\,|\,\mathbf{x})\right] = \mathbb{E}_{\mathbf{z}\sim q(\mathbf{z}|\mathbf{x})} \log p(\mathbf{x}\,|\,\mathbf{z}) - D\left[q(\mathbf{z}\,|\,\mathbf{x}) \,\|\, p(\mathbf{z})\right] \tag{2}$$

with (positive definite) Kullback-Leibler divergences $D[q \,\|\, p] \doteq \int d\mathbf{z}\, q(\mathbf{z}) \log\left[q(\mathbf{z})/p(\mathbf{z})\right]$ for generic distributions and their respective conditionals. Eq. (2) minimizes for $q(\mathbf{z}\,|\,\mathbf{x}) \to p(\mathbf{z}\,|\,\mathbf{x})$, leading to the definition of the evidence lower bound (ELBO) as:

$$\log p(\mathbf{x}) \geq \mathbb{E}_{\mathbf{z}\sim q_\varphi(\mathbf{z}|\mathbf{x})} \log p_\theta(\mathbf{x}\,|\,\mathbf{z}) - D\left[q_\varphi(\mathbf{z}\,|\,\mathbf{x}) \,\|\, p_\theta(\mathbf{z})\right] \tag{3}$$

where the distributions are now parameterized for use in DNNs. The $\{\theta, \varphi\}$ parameters are to be learned by e.g. Stochastic Gradient Descent (SGD) using the backpropagation algorithm.[1] The parameter estimation is global whereby parameters are obtained for the entire data set.

The language used in ML for the Bayes/Laplace-based expression $p_\theta(\mathbf{z}\,|\,\mathbf{x}) = p_\theta(\mathbf{z})\, p_\theta(\mathbf{x}\,|\,\mathbf{z}) / p(\mathbf{x})$ in words reads posterior = (prior X likelihood)/evidence.

Because a partition function $p(\mathbf{x}) = \int d\mathbf{z}\, p(\mathbf{x},\mathbf{z})$ is required the posterior is in general intractable. This motivates introduction of a stochastic encoder $q_{\varphi}(\mathbf{z}\,|\,\mathbf{x})$, also referred to as the inference model, to approximate the true but intractable posterior $p_{\theta}(\mathbf{z}\,|\,\mathbf{x})$ of the generative model. Thus, the goal is to find a method whereby $q_{\varphi}(\mathbf{z}\,|\,\mathbf{x}) \approx p_{\theta}(\mathbf{z}\,|\,\mathbf{x})$.

The ELBO approximation in Eq. (3) consists of a reconstruction term (first term on the RHS) which, if the only term, would amount to an autoencoder and a second, regularizer, term. The conventional choice of latent distribution is an i.i.d Gaussian $p_{\theta}(\mathbf{z}) \doteq q(\mathbf{z}) = \mathcal{N}_{\mathbf{z}}(\mathbf{0},\mathbf{I})$. This assumption, along with Gaussian models for the encoder and decoder, provides an analytical regularizer expression and a reconstruction term evaluable by (unbiased) MC estimation. Such simple choices do lead to issues with VAEs, leading to alternatives such as the InfoVAE [7] and $\beta$ - VAE methods. [8]

**3 Three new variational autoencoders**

Three variations on VAEs can be obtained by introducing two encoders: $V(\mathbf{z}\,|\,\mathbf{x})$ and $Y(\mathbf{z}\,|\,\mathbf{x})$ (or $U(\mathbf{z}\,|\,\mathbf{x})$) with parameters learnable using e.g. Stochastic Gradient Descent (SGD). Furthermore, we introduce an encoder distribution $W(\mathbf{z}\,|\,\mathbf{x})$ *not* to be treated as an intractable posterior but, rather, obtained once-and-for-all from the data by a method such as probabilistic Principle Component Analysis (P-PCA). We shall use different symbols from the above VAE/ELBO to put the two intractable posterior encoders, $V(\mathbf{z}\,|\,\mathbf{x})$ and $Y(\mathbf{z}\,|\,\mathbf{x})$, on the same footing. (For notational convenience in the following $D\left[E(\mathbf{z}\,|\,\mathbf{x}) \,\|\, F(\mathbf{z}\,|\,\mathbf{x})\right] \doteq D\left[E \,\|\, F\right]$).

The three methods are based on three connections between decoders and encoders using the Bayes/Laplace relation:

**(A)** $\tilde{V}(\mathbf{x}\,|\,\mathbf{z}) \doteq W(\mathbf{z}\,|\,\mathbf{x})\, p(\mathbf{x}) / q(\mathbf{z})$ so $W(\mathbf{z}\,|\,\mathbf{x})\, p(\mathbf{x}) = \tilde{V}(\mathbf{x}\,|\,\mathbf{z})\, q(\mathbf{z});\ q(\mathbf{z}) = \int d\mathbf{z}\, W(\mathbf{z}\,|\,\mathbf{x})\, p(\mathbf{x})$

**(B)** $\hat{V}(\mathbf{x}\,|\,\mathbf{z}) \doteq V(\mathbf{z}\,|\,\mathbf{x})\, p(\mathbf{x}) / q(\mathbf{z})$ so $V(\mathbf{z}\,|\,\mathbf{x})\, p(\mathbf{x}) = \hat{V}(\mathbf{x}\,|\,\mathbf{z})\, q(\mathbf{z});\quad q(\mathbf{z}) = \int d\mathbf{z}\, \hat{V}(\mathbf{z}\,|\,\mathbf{x})\, p(\mathbf{x})$ (4)

**(C)** $\bar{V}(\mathbf{x}\,|\,\mathbf{z}) \doteq U(\mathbf{z}\,|\,\mathbf{x})\, p(\mathbf{x}) / q(\mathbf{z})$ so $U(\mathbf{z}\,|\,\mathbf{x})\, p(\mathbf{x}) = \bar{V}(\mathbf{x}\,|\,\mathbf{z})\, q(\mathbf{z});\quad q(\mathbf{z}) = \int d\mathbf{z}\, \bar{V}(\mathbf{z}\,|\,\mathbf{x})\, p(\mathbf{x})$

with the last expressions for $q(\mathbf{z})$ emphasizing that there are two independent quantities. The $q(\mathbf{z})$ definitions, even if given in terms of a different decoder, respectively: $\tilde{V}(\mathbf{x}|\mathbf{z})$, $\hat{V}(\mathbf{x}|\mathbf{z})$ and $\bar{V}(\mathbf{x}|\mathbf{z})$, still induce an independent $q(\mathbf{z})$. In other words, $q(\mathbf{z})$ is transferred to optimization of the above three decoder $V(\mathbf{x}|\mathbf{z})$ forms.

**3.1 VAE_A**

The VAE expression written in the above notation is

$$\log p(\mathbf{x}) - D\left[V(\mathbf{z}|\mathbf{x}) \,\|\, W(\mathbf{z}|\mathbf{x})\right] = \mathbb{E}_{\mathbf{z}\sim V(\mathbf{z}|\mathbf{x})} \log \tilde{V}(\mathbf{x}|\mathbf{z}) - D\left[V(\mathbf{z}|\mathbf{x}) \,\|\, q(\mathbf{z})\right] \tag{5}$$

having used relation Eq. (4) (**A**), $\tilde{V}(\mathbf{x}|\mathbf{z}) \doteq W(\mathbf{z}|\mathbf{x})\, p(\mathbf{x})/q(\mathbf{z})$, to introduce the $\tilde{V}(\mathbf{x}|\mathbf{z})$ decoder version. Writing

$$\begin{aligned} D[V \,\|\, Y] &= \int d\mathbf{z}\, V(\mathbf{z}|\mathbf{x}) \log\left\{\left[\frac{V(\mathbf{z}|\mathbf{x})}{W(\mathbf{z}|\mathbf{x})}\right]\left[\frac{W(\mathbf{z}|\mathbf{x})}{Y(\mathbf{z}|\mathbf{x})}\right]\right\} \\ &= D[V \,\|\, W] - \int d\mathbf{z}\, V(\mathbf{z}|\mathbf{x}) \log\left[\frac{Y(\mathbf{z}|\mathbf{x})}{W(\mathbf{z}|\mathbf{x})}\right] \\ \therefore D[V \,\|\, W] &= D[V \,\|\, Y] + \int d\mathbf{z}\, V(\mathbf{z}|\mathbf{x}) \log\left[\frac{Y(\mathbf{z}|\mathbf{x})}{W(\mathbf{z}|\mathbf{x})}\right] \end{aligned} \tag{6}$$

and using it in Eq. (5) provides VAE_A:

$$\log p(\mathbf{x}) - D\left[V(\mathbf{z}|\mathbf{x}) \,\|\, Y(\mathbf{z}|\mathbf{x})\right] = \text{recon} - \text{regu} + \int d\mathbf{z}\, V(\mathbf{z}|\mathbf{x}) \log\left[\frac{Y(\mathbf{z}|\mathbf{x})}{W(\mathbf{z}|\mathbf{x})}\right] \quad (\text{VAE_A})$$

$$\begin{aligned} \text{recon} &\doteq \mathbb{E}_{\mathbf{z}\sim V(\mathbf{z}|\mathbf{x})} \log \tilde{V}(\mathbf{x}|\mathbf{z}); \\ \text{regu} &\doteq D\left[V(\mathbf{z}|\mathbf{x}) \,\|\, q(\mathbf{z})\right] \end{aligned} \tag{7}$$

with its definitions of, respectively, a reconstruction (recon) and regularizer (regu) term, in VAE terminology. This expression maintains a posterior $W(\mathbf{z}|\mathbf{x})$ to be modeled, and two intractable posteriors, $V_{\varphi_V}(\mathbf{z}|\mathbf{x})$ and $Y_{\varphi_Y}(\mathbf{z}|\mathbf{x})$, to be variationally optimized with a DNN and e.g. SGD.

In the "limits": $Y \to V$ or $Y \to W$, VAE_A goes back to the VAE in Eq. (2) because then there is only one encoder. If the $D[V \,\|\, Y]$ term is dropped the result is an ELBO version. However, because $W$ is a specified encoder it is formally *not* same as the standard VAE/ELBO approximation.

**3.2 VAE_B**

Another VAE version not incorporating a $W(\mathbf{z}|\mathbf{x})$ posterior proceeds via the identity

$$\begin{aligned} D[V \,\|\, Y] &= \int d\mathbf{z}\, V(\mathbf{z}|\mathbf{x}) \log\left[\frac{V(\mathbf{z}|\mathbf{x})U(\mathbf{z}|\mathbf{x})}{U(\mathbf{z}|\mathbf{x})Y(\mathbf{z}|\mathbf{x})}\right] \\ &= \int d\mathbf{z}\, V(\mathbf{z}|\mathbf{x}) \log\left[\frac{V(\mathbf{z}|\mathbf{x})}{U(\mathbf{z}|\mathbf{x})}\right] + \int d\mathbf{z}\, V(\mathbf{z}|\mathbf{x}) \log\left[\frac{U(\mathbf{z}|\mathbf{x})}{Y(\mathbf{z}|\mathbf{x})}\right] \\ \therefore \quad D[V \,\|\, Y] &= D[V \,\|\, U] + \int d\mathbf{z}\, V(\mathbf{z}|\mathbf{x}) \log\left[\frac{U(\mathbf{z}|\mathbf{x})}{Y(\mathbf{z}|\mathbf{x})}\right] \end{aligned} \tag{8}$$

where $U(\mathbf{z}|\mathbf{x})$ is another learned encoder. Now use as a decoder definition (**B**) in Eq. (4): $\hat{V}(\mathbf{x}|\mathbf{z}) \doteq V(\mathbf{z}|\mathbf{x})\, p(\mathbf{x})/q(\mathbf{z})$ and use this in $D[V \,\|\, U]$ to get

$$\begin{aligned} D[V \,\|\, U] &= \int d\mathbf{z}\, V(\mathbf{z}|\mathbf{x}) \log\left[\frac{\hat{V}(\mathbf{x}|\mathbf{z})q(\mathbf{z})}{U(\mathbf{z}|\mathbf{x})\, p(\mathbf{x})}\right] \\ &= \int d\mathbf{z}\, V(\mathbf{z}|\mathbf{x}) \log \hat{V}(\mathbf{x}|\mathbf{z}) + \int d\mathbf{z}\, V(\mathbf{z}|\mathbf{x}) \log\left[\frac{q(\mathbf{z})}{U(\mathbf{z}|\mathbf{x})}\right] - \log p(\mathbf{x}) \\ &\doteq \text{recon} + \int d\mathbf{z}\, V(\mathbf{z}|\mathbf{x}) \log\left[\frac{q(\mathbf{z})}{U(\mathbf{z}|\mathbf{x})}\right] - \log p(\mathbf{x}). \end{aligned} \tag{9}$$

Therefore, from this $D[V \,\|\, U]$ expression:

$$\log p(\mathbf{x}) + D[V \,\|\, U] = \text{recon} - \int d\mathbf{z}\, V(\mathbf{z}|\mathbf{x}) \log\left[\frac{U(\mathbf{z}|\mathbf{x})}{q(\mathbf{z})}\right] \quad (\text{VAE_B}) \tag{10}$$

This is an *upper* bound to $\log p(\mathbf{x})$; if $U \to V$ then $D[V \,\|\, U] \to 0$.

We will refer to the consequent approximation

$$\log p(\mathbf{x}) = \text{recon} - D\left[V(\mathbf{z}|\mathbf{x}) \,\|\, q(\mathbf{z})\right] \doteq \text{recon} - \text{regu} \tag{11}$$

as an Evidence *Upper* Bound (EUBO) approximation. It has a form similar to the standard ELBO but it provides an upper bound to $\log p(\mathbf{x})$.

In the VAE-ELBO in Eq. (2) one has to drop $D\left[q(\mathbf{z}|\mathbf{x}) \,\|\, p(\mathbf{z}|\mathbf{x})\right]$ because the intractable posterior $p(\mathbf{z}|\mathbf{x})$ is not available; in the EUBO, one does not have to drop $D[V \,\|\, U]$ because both $U$ and $V$ are parameterized as trial encoders.

Also note a symmetry of the VAE_B expression. Using similar algebra to the above,

$$D\left[V \parallel Y\right] = D\left[V \parallel U\right] + \int d\mathbf{z}\, V\left(\mathbf{z}\,|\,\mathbf{x}\right)\log\left[\frac{U\left(\mathbf{z}\,|\,\mathbf{x}\right)}{Y\left(\mathbf{z}\,|\,\mathbf{x}\right)}\right] \tag{12}$$

and use of Eq. (12) in Eq. (11) produces

$$\log p\left(\mathbf{x}\right) + D\left[V \parallel Y\right] = \mathrm{recon} - \int d\mathbf{z}\, V\left(\mathbf{z}\,|\,\mathbf{x}\right)\log\left[\frac{Y\left(\mathbf{z}\,|\,\mathbf{x}\right)}{q\left(\mathbf{z}\right)}\right]. \tag{13}$$

In the "limit" $Y \to V$, VAE_B goes back to the VAE-ELBO expression. However, as noted above, in EUBO $D\left[V \parallel Y\right]$ is evaluable because both $V$ and $Y$ are parameterized. This provides a method of "squinching down" to the correct result such that, when converged, $V \to Y \to W$, with, here, $W\left(\mathbf{z}\,|\,\mathbf{x}\right) = p\left(\mathbf{z}\,|\,\mathbf{x}\right)$, the true posterior. Thus, as the optimization proceeds, the EUBO and ELBO predictions for $W$ approach each other from, respectively, above and below. Of course, from the EUBO perspective, convergence can also be monitored from the difference between $V$ and $Y$. Said otherwise, if ELBO and EUBO approach each other then this does imply $V \to Y \to W$ because, if they approach each other, $D\left[V \parallel W\right] \to 0$.

It is worth analyzing the EUBO in some more detail. To this end, expand $\log Y$ around $\log V$ to second order in $\delta Y \doteq Y - V$:

$$\log Y = \log V + \log\left[1 + \delta Y/V\right] \approx \log V + \delta Y/V - \left(1/2\right)\left(\delta Y/V\right)^2. \tag{14}$$

Then, to this order,

$$D\left[V \parallel Y\right] \approx D^{(0)}\left[V \parallel Y\right] + D^{(1)}\left[V \parallel Y\right] + D^{(2)}\left[V \parallel Y\right] \tag{15}$$

with

$$\begin{aligned}
&D^{(0)}\left[V \parallel Y\right] = 0;\\
&D^{(1)}\left[V \parallel Y\right] = -\int d\mathbf{z}\left(V\left(\mathbf{z}\,|\,\mathbf{x}\right) - Y\left(\mathbf{z}\,|\,\mathbf{x}\right)\right) = 0;\\
&D^{(2)}\left[V \parallel Y\right] = \frac{1}{2}\int d\mathbf{z}\left(\frac{\delta Y^2\left(\mathbf{z}\,|\,\mathbf{x}\right)}{V\left(\mathbf{z}\,|\,\mathbf{x}\right)}\right) \geq 0.
\end{aligned} \tag{16}$$

To this order, $D\left[V \parallel Y\right] \approx D^{(2)}\left[V \parallel Y\right]$.

Also, expand $\int d\mathbf{z}\, V\left(\mathbf{z}\,|\,\mathbf{x}\right)\log Y\left(\mathbf{z}\,|\,\mathbf{x}\right)$ to this order as

$$-\int d\mathbf{z}\, V(\mathbf{z}\,|\,\mathbf{x})\log Y(\mathbf{z}\,|\,\mathbf{x}) \approx -\int d\mathbf{z}\left[\delta V(\mathbf{z}\,|\,\mathbf{x})\right] + \frac{1}{2}\int d\mathbf{z}\frac{\delta Y^2(\mathbf{z}\,|\,\mathbf{x})}{V(\mathbf{z}\,|\,\mathbf{x})} = D^{(2)}\left[V\,\|\,Y\right]. \quad (17)$$

Use of the above expansions in the EUBO yields

$$\log p(\mathbf{x}) + D^{(2)}\left[V\,\|\,Y\right] = \text{recon} + D^{(2)}\left[V\,\|\,Y\right] - \int d\mathbf{z}\, V(\mathbf{z}\,|\,\mathbf{x})\log\frac{V(\mathbf{z}\,|\,\mathbf{x})}{q(\mathbf{z})}. \quad (18)$$

Thus,

$$\log p(\mathbf{x}) = \text{recon} - \text{regu}. \quad (19)$$

Therefore, $\text{EUBO}^{(2)}$ = ELBO and one regains the VAE-ELBO form.

In Appendix A another version of VAE_B using a Jensen-Shannon versus a Kullback-Leibler divergence is given.

### 3.3 VAE_C

As a last VAE version, use the Eq. (4) (**C**) decoder definition $\bar{V}(\mathbf{x}\,|\,\mathbf{z}) \doteq U(\mathbf{z}\,|\,\mathbf{x})\, p(\mathbf{x})/q(\mathbf{z})$ to get, using manipulations similar to what has been done above,

$$\log p(\mathbf{x}) - D\left[V\,\|\,U\right] = \text{recon} - \int d\mathbf{z}\, V(\mathbf{z}\,|\,\mathbf{x})\log\left[\frac{V(\mathbf{z}\,|\,\mathbf{x})}{q(\mathbf{z})}\right] \doteq \text{recon - regu} \quad (\text{VAE_C}) \quad (20)$$

This has the form of a VAE, again with two learnable encoders, but convergence means $V \to U$ minimizing $\log p(\mathbf{x})$, in contrast to the VAE-ELBO.

## 4 VAE_A explicitly

VAE_A incorporates two learnable encoders, $V_{\varphi_V}(\mathbf{z}\,|\,\mathbf{x})$ and $Y_{\varphi_Y}(\mathbf{z}\,|\,\mathbf{x})$, and another, fixed encoder $W(\mathbf{z}\,|\,\mathbf{x})$. To make this VAE explicit requires a model for $W$. As PCA is one method to produce a latent space representation we will use a generalization of it known as Probabilistic PCA (P-PCA) introduced by Roweis [9], and use it as a model for this intractable posterior. PCA has a long history in data reduction/representation.[10] It certainly has deficiencies but its P-PCA generalization fits in well with VAE schemes as noted below.

In our notation, Roweis writes the connection between latent and data space as:

$$\mathbf{x}(N_x) = \mathbf{C}^R(N_x \times N_z)\bullet\mathbf{z}(N_z) + \mathbf{v}(N_x);\;\; \mathbf{z} \sim \mathcal{N}(0,\mathbf{I}),\, \mathbf{v} \sim \mathcal{N}(0,\mathbf{R})\;\; \langle\mathbf{v}\mathbf{v}\rangle = R\mathbf{I} \quad (21)$$

where $\mathbf{z} \sim \prod_{k=1}^{N_z}\mathcal{N}(0,1)$ and $\mathbf{v} \sim \prod_{k=1}^{N_x}\mathcal{N}(0,R)$ are i.i.d. Gaussians.

The expression is pointwise in the $\mathbf{x}_i \left(i=1,2,...,N\right)$ data. $\mathbf{C}^R\left(N_x \times N_z\right)$ is the Roweis covariance matrix. How this PCA-projected covariance matrix $\mathbf{C}^R\left(N_x \times N_z\right)$ is obtained from the data covariance matrix $\mathbf{C}\left(N_x \times N_x\right) \doteq \frac{1}{N}\sum_{i=1}^{N}\mathbf{x}_i^T\mathbf{x}_i$ of dimension $N_x \times N_x$ will be given later. The strong assumption of i.i.d. Gaussian-distributed latents modeling the data is consonant with VAE implementations.

For this model,

$$p(\mathbf{x}) = \mathcal{N}_x\left(\mu_x, \Sigma_x\right) \text{ with } \mu_x = \mathbf{0};\ \Sigma_x = \mathbf{C}^R \bullet \mathbf{C}^{R,T} + \mathbf{R}. \tag{22}$$

The form follows from: 1) the change of variables rule in a probability distribution – if $\mathbf{z}$ is a centered Gaussian with $\Sigma_z = \mathbf{I}$ then $\mathbf{x} = \mathbf{C}^R \bullet \mathbf{z}$ is a centered Gaussian with $\Sigma_x = \mathbf{C}^R \bullet \mathbf{C}^{R,T}$ and 2) linearly additive noise adds on a $\Sigma_{\mathrm{v}} \doteq R\mathbf{I}$ covariance term.

For fixed $\mathbf{z}$, the source of randomness comes from $\mathbf{v}$ with its covariance $\Sigma_{\mathrm{v}} = R\mathbf{I}$ and the mean $\mu_x$ follows from the specification $\mathbf{z} \Rightarrow \mathbf{x} = \mathbf{C}^R \bullet \mathbf{z}$. Thus, the likelihood $p(\mathbf{x} \mid \mathbf{z})$ is

$$p(\mathbf{x} \mid \mathbf{z}) = \mathcal{N}_x\left(\mathbf{C}^R \bullet \mathbf{z}, \mathbf{R}\right). \tag{23}$$

Combining Eqs. (21)-(23), the (tractable) model posterior is

$$p(\mathbf{z} \mid \mathbf{x}) = p(\mathbf{x} \mid \mathbf{z})\, p(\mathbf{z}) / p(\mathbf{x}) = \mathcal{N}_{x|z}\left(\mathbf{C}^R \bullet \mathbf{z}, \mathbf{R}\right) \mathcal{N}_z\left(\mathbf{0}, \mathbf{I}\right) / \mathcal{N}_x\left(\mathbf{0}, \mathbf{C}^R \bullet \mathbf{C}^{R,T} + \mathbf{R}\right). \tag{24}$$

Doing the Gaussian arithmetic yields:

$$p(\mathbf{z} \mid \mathbf{x}) \doteq W(\mathbf{z} \mid \mathbf{x}) = \mathcal{N}_z\left(\beta\mathbf{x}, \mathbf{I} - \beta\mathbf{C}^R\right);\ \beta \doteq \mathbf{C}^{R,T}\left(\mathbf{C}^R \bullet \mathbf{C}^{R,T} + \mathbf{R}\right)^{-1}. \tag{25}$$

The expected values of the latent variables depend on the additive noise with covariance $\Sigma_{\mathrm{v}} = R\mathbf{I}$ via $\beta$. This completely specifies the P-PCA model.

Roweis evaluated the P-PCA model with an EM algorithm though, as noted there [9], construction of the covariance matrix $\mathbf{C}^R\left(N_x \times N_z\right)$ connecting data and latent spaces is obtained by standard PCA means. In the PCA-limit, R→0, one sensibly gets from, respectively, Eqs. (23) and (25), the noiseless limits

$$p(\mathbf{x} \mid \mathbf{z}) = \delta\left(\mathbf{x} - \mathbf{C}^R \bullet \mathbf{z}\right) \text{ and } p(\mathbf{z} \mid \mathbf{x}) = \delta\left(\mathbf{z} - \left(\mathbf{C}^{R,T} \bullet \mathbf{C}^R\right)^{-1} \bullet \mathbf{C}^{R,T} \bullet \mathbf{x}\right). \tag{26}$$

This latter limit uses

$$\lim_{R\to 0}\left(\mathbf{I}-\beta\mathbf{C}^R\right)=\mathbf{I}-\mathbf{C}^{R,T}\bullet\left(\mathbf{C}^R\bullet\mathbf{C}^{R,T}\right)^{-1}\bullet\mathbf{C}^R=\mathbf{I}-\left(\mathbf{C}^{R,T}\bullet\mathbf{C}^R\right)^{-1}\bullet\mathbf{C}^{R,T}\bullet\mathbf{C}^R$$
$$=\mathbf{I}-\mathbf{C}^{R,T}\bullet\left(\mathbf{C}^{R,T}\right)^{-1}\bullet\left(\mathbf{C}^R\right)^{-1}\bullet\mathbf{C}^R=\mathbf{I}-\mathbf{I}=\mathbf{0}.$$

The second equality in the above chain relies on the equality $\mathbf{C}^{R,T}\bullet\left(\mathbf{C}^R\bullet\mathbf{C}^{R,T}\right)^{-1}=\left(\mathbf{C}^{R,T}\bullet\mathbf{C}^R\right)^{-1}\bullet\mathbf{C}^{R,T}$, permitting projection of the null space of the non-invertible $\left(\mathbf{C}^R\bullet\mathbf{C}^{R,T}\right)^{-1}$ onto the finite $\mathbf{C}^{R,T}$. In the noiseless, PCA, limit, the posterior collapses to a single point, as the covariance $\mathbf{I}-\beta\mathbf{C}^R\to 0$. Thus, $\mathbf{z}=\left(\mathbf{C}^{R,T}\bullet\mathbf{C}^R\right)^{-1}\bullet\mathbf{C}^{R,T}\bullet\mathbf{x}$

The Roweis covariance matrix $\mathbf{C}^R\left(N_x\times N_z\right)$ can be obtained from the $\mathbf{C}\left(N_x\times N_x\right)$ data covariance matrix in this PCA subspace via $\mathbf{C}^R\left(N_x\times N_z\right)\doteq\left(\mathbf{m}_1,\ldots,\mathbf{m}_{N_z}\right)$, where the $\mathbf{m}_k\ \left(k=1,2,\ldots,N_z\right)$ are the $N_z$ eigenvalue-descending-ordered PCA eigenvectors obtained from the PCA decomposition

$$\mathbf{C}\left(N_x\times N_x\right)\bullet\mathbf{m}_k=\mu_k\mathbf{m}_k\ \left(k=1,2,\ldots,N_x\right). \tag{27}$$

With the P-PCA model for $W\left(\mathbf{z}\,|\,\mathbf{x}\right)$ and conventional Gaussian models for the learnable encoders $V_{\varphi_V}\left(\mathbf{z}\,|\,\mathbf{x}\right)$ and $Y_{\varphi_Y}\left(\mathbf{z}\,|\,\mathbf{x}\right)$, explicit forms for the required integrals can be obtained. The “new term” in VAE_A can be written as

$$\begin{aligned}&\int d\mathbf{z}V_{\varphi_V}\left(\mathbf{z}\,|\,\mathbf{x}\right)\log\left[\frac{Y_{\varphi_Y}\left(\mathbf{z}\,|\,\mathbf{x}\right)}{W\left(\mathbf{z}\,|\,\mathbf{x}\right)}\right]\\&=\int d\mathbf{z}V_{\varphi_V}\left(\mathbf{z}\,|\,\mathbf{x}\right)\log Y_{\varphi_Y}\left(\mathbf{z}\,|\,\mathbf{x}\right)-\int d\mathbf{z}V_{\varphi_V}\left(\mathbf{z}\,|\,\mathbf{x}\right)\log W\left(\mathbf{z}\,|\,\mathbf{x}\right).\end{aligned} \tag{28}$$

Also

$$\begin{aligned}&\int d\mathbf{z}V_{\varphi_V}\left(\mathbf{z}\,|\,\mathbf{x}\right)\log Y_{\varphi_Y}\left(\mathbf{z}\,|\,\mathbf{x}\right)\\&=\int d\mathbf{z}V_{\varphi_V}\left(\mathbf{z}\,|\,\mathbf{x}\right)\log V_{\varphi_V}\left(\mathbf{z}\,|\,\mathbf{x}\right)-\int d\mathbf{z}V_{\varphi_V}\left(\mathbf{z}\,|\,\mathbf{x}\right)\log\left[\frac{V_{\varphi_V}\left(\mathbf{z}\,|\,\mathbf{x}\right)}{Y_{\varphi_Y}\left(\mathbf{z}\,|\,\mathbf{x}\right)}\right]\\&=-H\left(V_{\varphi_V}\left(\mathbf{z}\,|\,\mathbf{x}\right)\right)-D\left[V_{\varphi_V}\left(\mathbf{z}\,|\,\mathbf{x}\right)\|\,Y_{\varphi_Y}\left(\mathbf{z}\,|\,\mathbf{x}\right)\right]\end{aligned} \tag{29}$$

and

$$
\begin{aligned}
&\int d\mathbf{z} V_{\varphi_V}(\mathbf{z}\,|\,\mathbf{x}) \log W(\mathbf{z}\,|\,\mathbf{x}) \\
&= \int d\mathbf{z} V_{\varphi_V}(\mathbf{z}\,|\,\mathbf{x}) \log V_{\theta_V}(\mathbf{z}\,|\,\mathbf{x}) - \int d\mathbf{z} V_{\varphi_V}(\mathbf{z}\,|\,\mathbf{x}) \log\left[\frac{V_{\varphi_V}(\mathbf{z}\,|\,\mathbf{x})}{W(\mathbf{z}\,|\,\mathbf{x})}\right] \\
&= -H\left(V_{\varphi_V}(\mathbf{z}\,|\,\mathbf{x})\right) - D\left[V_{\varphi_V}(\mathbf{z}\,|\,\mathbf{x}) \,\|\, W(\mathbf{z}\,|\,\mathbf{x})\right].
\end{aligned} \tag{30}
$$

Then

$$
\begin{aligned}
&\int d\mathbf{z} V_{\varphi_V}(\mathbf{z}\,|\,\mathbf{x}) \log\left[\frac{Y_{\varphi_Y}(\mathbf{z}\,|\,\mathbf{x})}{W(\mathbf{z}\,|\,\mathbf{x})}\right] \\
&= -H\left(V_{\varphi_V}(\mathbf{z}\,|\,\mathbf{x})\right) - D\left[V_{\varphi_V}(\mathbf{z}\,|\,\mathbf{x}) \,\|\, Y_{\varphi_Y}(\mathbf{z}\,|\,\mathbf{x})\right] \\
&\quad + H\left(V_{\varphi_V}(\mathbf{z}\,|\,\mathbf{x})\right) + D\left[V_{\varphi_V}(\mathbf{z}\,|\,\mathbf{x}) \,\|\, W(\mathbf{z}\,|\,\mathbf{x})\right] \\
&= D\left[V_{\varphi_V}(\mathbf{z}\,|\,\mathbf{x}) \,\|\, W(\mathbf{z}\,|\,\mathbf{x})\right] - D\left[V_{\varphi_V}(\mathbf{z}\,|\,\mathbf{x}) \,\|\, Y_{\varphi_Y}(\mathbf{z}\,|\,\mathbf{x})\right].
\end{aligned} \tag{31}
$$

Using i.i.d. Gaussians for these conditionals, as in the simplest-form ELBO-based calculations, the requisite divergences are:

$$
D\left[V_{\varphi_V}(\mathbf{z}\,|\,\mathbf{x}) \,\|\, Y_{\varphi_Y}(\mathbf{z}\,|\,\mathbf{x})\right] = \sum_{k=1}^{N_z} D_k\left[V_{\varphi_V}(z_k\,|\,\mathbf{x}) \,\|\, Y_{\varphi_Y}(z_k\,|\,\mathbf{x})\right] \tag{32}
$$

with k'th contribution

$$
D_k\left[V_{\varphi_V}(z_k\,|\,\mathbf{x}) \,\|\, Y_{\varphi_Y}(z_k\,|\,\mathbf{x})\right] = \frac{1}{2}\left\{\log\left(\frac{\Sigma^Y_{z_k}(\mathbf{x})}{\Sigma^V_{z_k}(\mathbf{x})}\right) + \frac{\Sigma^V_{z_k}(\mathbf{x})}{\Sigma^Y_{z_k}(\mathbf{x})} - 1 + \frac{\left(\mu^Y_{z_k}(\mathbf{x}) - \mu^V_{z_k}(\mathbf{x})\right)^2}{\Sigma^Y_{z_k}(\mathbf{x})}\right\}. \tag{33}
$$

Because the P-PCA form for $W(\mathbf{z}\,|\,\mathbf{x})$ is also a Gaussian, specialization of the generic Gaussian-Gaussian divergence expression [11] can be used to obtain

$$
D\left[V_{\varphi_V}(\mathbf{z}\,|\,\mathbf{x}) \,\|\, W(\mathbf{z}\,|\,\mathbf{x})\right] = \frac{1}{2}\left\{\begin{aligned}&\log\left(\frac{|\mathbf{1} - \beta\bullet\mathbf{C}|}{|\mathbf{1}\Sigma^V_z(\mathbf{x})|}\right) + \mathrm{Tr}\left[(\mathbf{1} - \beta\bullet\mathbf{C})^{-1}\bullet\mathbf{1}\Sigma^V_z(\mathbf{x})\right] - N_z \\ &+ \left(\beta\bullet\mathbf{x} - \mu^V_z(\mathbf{x})\right)^T \bullet (\mathbf{1} - \beta\bullet\mathbf{C})^{-1} \bullet \left(\beta\bullet\mathbf{x} - \mu^V_z(\mathbf{x})\right)\end{aligned}\right\}. \tag{34}
$$

Expressions for the $\mathrm{recon} \doteq \mathbb{E}_{\mathbf{z}\sim V(\mathbf{z}|\mathbf{x})} \log \tilde{V}(\mathbf{x}\,|\,\mathbf{z})$ and $\mathrm{regu} \doteq D\left[V(\mathbf{z}\,|\,\mathbf{x}) \,\|\, q(\mathbf{z})\right]$ follow from the conventional VAE-ELBO approach where, with again the simplest i.i.d Gaussian

encoder/decoder/prior forms, e.g. encoder: $V_{\varphi_V}\left(z_k \mid \mathbf{x}\right)=\mathcal{N}\left(z_k \mid \mu_{z_k}^V\left(\mathbf{x}\right),\Sigma_{z_k}^V\left(\mathbf{x}\right)\right)$, SGD on minibatchs of data to update the encoder and decoder parameters can be used.

Eq. (34) nominally requires inversion of an $N_x \times N_x$ matrix, which would typically be impractical. In Appendix B we show that evaluation of $D\left[V_{\varphi_V}\left(\mathbf{z} \mid \mathbf{x}\right) \| W\left(\mathbf{z} \mid \mathbf{x}\right)\right]$ can be obtained from the Singular Value Decomposition (SVD) of the data covariance matrix that is trivial to obtain by standard numerical methods. It evaluates to:

$$
\begin{aligned}
& D\left[V_{\varphi_V}\left(\mathbf{z} \mid \mathbf{x}\right) \| W\left(\mathbf{z} \mid \mathbf{x}\right)\right] \\
& =\frac{1}{2}\left\{\begin{array}{l}
\sum_{l=1}^{N_z}\left\{\left[\sum_{j=1}^{N_z}\left(1+\lambda_j^2\right) V_{lj}\left(V^T\right)_{jl}\right] \Sigma_{z,l}\left(\mathbf{x}\right)-1-\log\left(1+\lambda_l^2\right) \Sigma_{z,l}\left(\mathbf{x}\right)\right\} \\
+\sum_{l=1}^{N_z}\left[\left(\overline{\mathbf{x}}-\bar{\mu}_z^V\left(\mathbf{x}\right)\right)^T \bullet \mathbf{u}_l\right]\left(H_d\right)_{ll}\left[\mathbf{u}_l \bullet\left(\overline{\mathbf{x}}-\bar{\mu}_z^V\left(\mathbf{x}\right)\right)\right]
\end{array}\right\}
\end{aligned} \tag{35}
$$

with $\mathbf{H}_d$ a diagonal matrix with elements

$$
\left(H_d\right)_{ll} \doteq \lambda_l^2 / \left(1+\lambda_l^2\right) \tag{36}
$$

and

$$
\overline{\mathbf{x}} \doteq \mathbf{x}/\sigma; \quad \bar{\boldsymbol{\mu}}_z^V\left(\mathbf{x}\right) \doteq \bar{\beta}^{-1} \boldsymbol{\mu}_z^V\left(\mathbf{x}\right)/\sigma=\left\{\left[\mathbf{I}_x+\mathbf{UL}\left(\mathbf{UL}\right)^T\right] \mathbf{UL}^{-1} \mathbf{V}^T\right\}\left(\boldsymbol{\mu}_z^V\left(\mathbf{x}\right)/\sigma\right). \tag{37}
$$

$\mathbf{C}^R/\sigma=\mathbf{ULV}^T$ is the singular value decomposition of the Roweis Covariance matrix. The SVD provides matrices $\mathbf{U}\left(N_x \times N_z\right)/\mathbf{V}\left(N_z \times N_z\right)$ of the left/right eigenvectors $\mathbf{u}_l / \mathbf{v}_l$ and descending-size-ordered eigenvalues $\lambda_l$ of the diagonal matrix $\mathbf{L}\left(N_z \times N_z\right)$. Using the first $N_z < N_x$ of them provides the desired PCA representation of $\mathbf{C}^R$. The means $\mu_z^V\left(\mathbf{x}\right)$ and variances $\Sigma_{z_l}^V\left(\mathbf{x}\right)$ of the encoder $V_{\varphi_V}\left(\mathbf{z} \mid \mathbf{x}\right)$ are to be learned.

**5. Conclusions**

Of these three variations on variational autoencoders two, (VAE_B, Eq. (10) and VAE_C, Eq. (20)), have two intractable posteriors, necessitating introduction of two variationally optimized encoders, along with one optimized decoder. In contrast with the VAE, where the $D\left[q_\varphi\left(\mathbf{z} \mid \mathbf{x}\right) \| p_\theta\left(\mathbf{z} \mid \mathbf{x}\right)\right]$ governs accuracy in terms of the estimation of the true encoder, VAE_B's

accuracy is measured by $D\left[V(\mathbf{z}\,|\,\mathbf{x})\,||\,Y(\mathbf{z}\,|\,\mathbf{x})\right]$. VAE_B provides an evidence upper bound (EUBO) on the accuracy and, in this way, by monitoring the progress of both encoders, convergence can be established. If both VAE and VAE_B are evaluated, the difference of their respective approaches from lower and higher values also provides a good measure of convergence. As shown in Section 3.2, to second order in their difference, $Y(\mathbf{z}\,|\,\mathbf{x})-V(\mathbf{z}\,|\,\mathbf{x})$, the VAE-ELBO is regained. The difference between VAE_B in Eq. (10) and VAE_C in Eq. (20) is their different regularizer forms.

VAE_A incorporates, in contrast with VAE_B and VAE_C, in addition to the two optimizable encoders and a decoder, an approximation to the true intractable posterior, $W(\mathbf{z}\,|\,\mathbf{x})$. A robust model for $W(\mathbf{z}\,|\,\mathbf{x})$ can be obtained by using P-PCA. The new term in Eq. (30) , $D\left[V_{\varphi_V}(\mathbf{z}\,|\,\mathbf{x})\,||\,W(\mathbf{z}\,|\,\mathbf{x})\right]$ , is evaluable using SVD of the Roweis covariance matrix. The reconstruction term can be obtained by VAE-based Monte Carlo estimation. Thus, the additional learning cost to VAE_A is about the same as that for VAE_B and VAE_C.

VAE_A with its assumption of a factored Gaussian prior $p(\mathbf{z})$ for $W(\mathbf{z}\,|\,\mathbf{x})$, as also used in implementations of the standard VAE, removes a source of inconsistency. This is an important point as the choice of a prior conditions the rest of the algorithm; i.e. $p(\mathbf{x})=\int p(\mathbf{x}\,|\,\mathbf{z})\,p(\mathbf{z})$ in both learning and generative phases.

**Appendix A**

The Jensen-Shannon divergence $JSD \doteq J$ is a symmetrized version of the Kullback-Leibler divergence $KLD \doteq D$:

$$J[U\,||\,V] \doteq \frac{1}{2}D[U\,||\,M]+\frac{1}{2}D[V\,||\,M];\;\; M \doteq \frac{1}{2}(U+V). \tag{A.1}$$

It is therefore suited to VAE_B with two encoders and their respective decoders treated on the same footing. Connect the encoders to decoders via

$$\hat{L}(\mathbf{x}\,|\,\mathbf{z})=L(\mathbf{z}\,|\,\mathbf{x})\,p(\mathbf{x})/q(\mathbf{z})\;\; L\in\{U,V\} \tag{A.2}$$

such that

$$D[L \| M] = \int d\mathbf{z}\, L(\mathbf{z}|\mathbf{x}) \log\left[\frac{\hat{L}(\mathbf{x}|\mathbf{z})q(\mathbf{z})}{p(\mathbf{x})} \frac{1}{M(\mathbf{z}|\mathbf{x})}\right] \quad L \in \{U, V\}. \tag{A.3}$$

Combine terms as

$$\begin{aligned} J[V \| U] &= \frac{1}{2}\left\{\int d\mathbf{z}\, V(\mathbf{z}|\mathbf{x}) \log \hat{V}(\mathbf{x}|\mathbf{z}) + \int d\mathbf{z}\, U(\mathbf{z}|\mathbf{x}) \log \hat{U}(\mathbf{x}|\mathbf{z})\right\} \\ &+ \int d\mathbf{z} M(\mathbf{z}|\mathbf{x}) \log\left[\frac{q(\mathbf{z})}{M(\mathbf{z}|\mathbf{x})}\right] - \int d\mathbf{z} M(\mathbf{z}|\mathbf{x}) \log p(\mathbf{x}) \\ &= \int d\mathbf{z} M(\mathbf{z}|\mathbf{x}) \log\left[\frac{q(\mathbf{z})}{M(\mathbf{z}|\mathbf{x})}\right] - \log p(\mathbf{x}) \end{aligned} \tag{A.4}$$

with the last equality following from

$$\int d\mathbf{z} M(\mathbf{z}|\mathbf{x}) \log p(\mathbf{x}) = \frac{1}{2}\int d\mathbf{z}\left[U(\mathbf{z}|\mathbf{x}) + V(\mathbf{z}|\mathbf{x})\right] \log p(\mathbf{x}) = \log p(\mathbf{x}). \tag{A.5}$$

Rearranging provides another, symmetric, version of the VAE_B

$$\log p(\mathbf{x}) + J[V \| U] = \frac{1}{2}\left\{\text{recon}_V + \text{recon}_U\right\} - D[M \| q] \tag{A.6}$$

with its EUBO form.

**Appendix B**

The divergence term that involves the P-PCA model distribution $W(\mathbf{z}|\mathbf{x})$ can be expressed in terms of the SVD of the Roweis covariance matrix $\mathbf{C}^R$. Write for notational convenience

$$D\left[V_{\varphi_V}(\mathbf{z}|\mathbf{x}) \| W(\mathbf{z}|\mathbf{x})\right] = D_1\left[V_{\varphi_V}(z|x) \| W(z|x)\right] + D_2\left[V_{\varphi_V}(z|x) \| W(z|x)\right] \tag{B.1}$$

with

$$\begin{aligned} &D_1\left[V_{\varphi_V}(z|x) \| W(z|x)\right] \\ &\doteq \frac{1}{2}\left\{\log\left[\left|\left(I_z - \beta C^R\right)\right| \Big/ \left|I_z \Sigma_z^V(x)\right|\right] + \mathrm{Tr}\left[\left(I_z - \beta C^R\right)^{-1} I_z \Sigma_z^V(x)\right] - N_z\right\} \end{aligned} \tag{B.2}$$

$$D_2\left[V_{\varphi_V}(z|x) \| W(z|x)\right] \doteq \frac{1}{2}\left(\beta x - \mu_z^V(x)\right)^T \left(I_z - \beta C^R\right)^{-1} \left(\beta x - \mu_z^V(x)\right) \tag{B.3}$$

where

$$\beta \doteq \left(C^R\right)^T \left[C^R \left(C^R\right)^T + \sigma^2 I_x\right]^{-1}. \tag{B.4}$$

To evaluate the terms contributing to $D\left[V_{\varphi_V}(\mathbf{z}|\mathbf{x})\|W(\mathbf{z}|\mathbf{x})\right]$, it proves convenient to define the latent-space-dimension matrix

$$J(N_z \times N_z) \doteq \left[I_z - \beta C^R\right]^{-1} = \left[I_z - \left(C^R\right)^T \left[C^R\left(C^R\right)^T + \sigma^2 I_x\right]^{-1} C^R\right]^{-1}$$
$$\doteq \left[I_z - \Lambda^T\left(\Lambda\Lambda^T + I_x\right)^{-1}\Lambda\right]^{-1}; \quad \Lambda \doteq C^R/\sigma. \tag{B.5}$$

Note that $\Lambda(N_x \times N_z)$ and $\Lambda^T(N_z \times N_x)$. To eliminate inversion of $\left[\Lambda\Lambda^T + I_x\right](N_x \times N_x)$, use the Woodbury matrix identity [12]

$$A^{-1} - A^{-1}U\left(C^{-1} + VA^{-1}U\right)^{-1}VA^{-1} = \left(A + UCV\right)^{-1}. \tag{B.6}$$

Setting: $A = I_z; C = I_x; V = \Lambda; U = \Lambda^T$ and noting $I_z^{-1} = I_z$, $I_x^{-1} = I_x$,

$$J^{-1} = I_z - \Lambda^T\left(I_x + \Lambda\Lambda^T\right)^{-1}\Lambda = \left(I_z + \Lambda^T\Lambda\right)^{-1}$$
$$\therefore J = \left[I_z - \beta C^R\right]^{-1} = I_z + \Lambda^T\Lambda\,. \tag{B.7}$$

Eq. (B.7) a) expresses the desired inverse in the $\Lambda^T\Lambda(N_z \times N_z)$ latent space and b) analytically inverts the required inverse.

To obtain more explicit expressions, use the SVD representation of the Roweis covariance matrix: $C^R/\sigma \doteq \Lambda = ULV^T$ , where $L(N_z \times N_z)$ is a diagonal matrix of the first $N_z \le N_x$ descending-ordered eigenvalues $\lambda_l$ $(l = 1,2,..,N_z)$, $U(N_x \times N_z)$ is a column orthonormal matrix and $V(N_z \times N_z)$ is a row/column orthonormal matrix such that

$$U^TU = V^TV = VV^T = I_z \text{ and } V^{-1} = V^T \text{ and } V = \left(V^{-1}\right)^T = \left(V^T\right)^{-1}. \tag{B.8}$$

Note that with Eq.(B.8)

$$\Lambda^T\Lambda = \left(ULV^T\right)^T\left(ULV^T\right) = \left(VLU^T\right)\left(ULV^T\right) = VL^2V^T\,. \tag{B.9}$$

For the trace term in Eq. (B.2) use of Eqs. (B.7) and (B.9) provides

$$\mathrm{TR} \doteq \mathrm{Tr}\left[\left(I_z - \beta C^R\right)^{-1} I_z \Sigma_z^V(x)\right] = \mathrm{Tr}\left[\left(I_z + \Lambda^T\Lambda\right)\Sigma_z^V(x)\right]$$
$$= \mathrm{Tr}\left[I_z + VL^2V^T\right]\Sigma_z^V(x) = \sum_{l=1}^{N_z}\left[1 + \sum_{j=1} V_{lj}\lambda_j^2\left(V^T\right)_{jl}\right]\Sigma_{z,l}^V(x). \tag{B.10}$$

For the log determinant term in Eq. (B.2), from Eq. (B.7) and noting that $|AB| = |A||B|$ and $|A|^{-1} = |A^{-1}|$,

$$\begin{aligned}\mathrm{LT} &\doteq \log\left[\left|\left(I_z - \beta C^R\right)\right| \Big/ \left|I_z \Sigma_z^V(x)\right|\right] = -\log\left[\left|\left(I_z - \beta C^R\right)\right|^{-1} \left|I_z \Sigma_z^V(x)\right|\right] \\ &= -\log\left[\left(I_z + \Lambda^T \Lambda\right)\left|I_z \Sigma_z^V(x)\right|\right].\end{aligned} \tag{B.11}$$

Using the SVD and Eq. (B.9)

$$\begin{aligned}\left|\left(I_z + \Lambda^T \Lambda\right)\right| &= \left|\left(I_z + VL^2V^T\right)\right| = \left|\left(VI_zV^T + VL^2V^T\right)\right| = \left|V\left(I_z + L^2\right)V^T\right| \\ &= |V|\left|\left(I_z + L^2\right)\right|\left|V^T\right| = (\pm 1)^2 \left|\left(I_z + L^2\right)\right| = \prod_{l=1}^{N_z}\left(1 + \lambda_l^2\right).\end{aligned} \tag{B.12}$$

Therefore, the log determinant term in Eq. (B.2) is

$$\begin{aligned}\mathrm{LT} &= \log\left[\left|\left(I_z - \beta C^R\right)\right| \Big/ \left|I_z \Sigma_z^V(x)\right|\right] = -\log\prod_{l=1}^{N_z}\left(1 + \lambda_l^2\right) - \log\prod_{l=1}^{N_z}\left(\Sigma_{z,l}(x)\right) \\ &= -\sum_{l=1}^{N_z}\log\left[\left(1 + \lambda_l^2\right)\Sigma_{z,l}(x)\right].\end{aligned} \tag{B.13}$$

Combining the trace and log-determinant terms and subtracting $N_z$:

$$\begin{aligned}D_1&\left[V_{\varphi_V}(z \mid x) \,\|\, W(z \mid x)\right] = \mathrm{TR} + \mathrm{LT} - N_z \\ &= \sum_{l=1}^{N_z}\left[\Sigma_{z,l}(x) + \sum_{j=1} V_{lj}\lambda_j^2 \left(V^T\right)_{jl} \Sigma_{z,l}(x)\right] - \sum_{l=1}^{N_z}\log\left(1 + \lambda_l^2\right)\Sigma_{z,l}(x) - \sum_{l=1}^{N_z} 1 \\ &= \sum_{l=1}^{N_z}\left\{\left[\sum_{j=1}\left(1 + \lambda_j^2\right)V_{lj}\left(V^T\right)_{jl}\right]\Sigma_{z,l}(x) - 1 - \log\left(1 + \lambda_l^2\right)\Sigma_{z,l}(x)\right\}\end{aligned} \tag{B.14}$$

where we have used orthonormality of the *V* matrix: $\left(VV^T\right)_{ll} = 1$ in the first term of the last line in Eq. (B.14).

It is not obvious from Eq. (B.14) that this combination of contributions to the divergence is non-negative. Appendix C shows that the generic version of $\mathrm{TR} + \mathrm{LT} - N_z$, there denoted as $Y - N_z \doteq \log\left(|\Sigma_J||\Sigma_I|^{-1}\right) + \mathrm{Tr}\left(\Sigma_J^{-1}\Sigma_I\right) - N_z$, is indeed positive semidefinite.

Turning to the scalar term, Eq. (B.3), make a dimensionless version using

$$\beta x-\mu_z^V\left(x\right)\doteq\bar{\beta}\left(\bar{x}-\bar{\mu}_z^V\left(x\right)\right);\quad \bar{x}\doteq x/\sigma;\quad \bar{\beta}\doteq\beta\sigma\quad \bar{\mu}_z^V\left(x\right)\doteq\mu_z^V\left(x\right)/\beta\sigma\,. \tag{B.15}$$

Note that the first equality relies on the facts that $\beta$, defined in Eq. (B.4), is *fixed* and $\mu_z^V\left(x\right)$ will be *learned*.

Then,

$$\left(\beta x-\mu_z^V\left(x\right)\right)^T\left(I_x-\beta C^R\right)^{-1}\left(\beta x-\mu_z^V\left(x\right)\right)\rightarrow\left(\bar{x}-\bar{\mu}_z^V\left(x\right)\right)^T S\left(\bar{x}-\bar{\mu}_z^V\left(x\right)\right) \tag{B.16}$$

with

$$S\doteq\bar{\beta}^T\left(I_x-\beta C^R\right)^{-1}\bar{\beta}\doteq(1)(2)(3) \tag{B.17}$$

where we define from Eq. (B.7) for $J$:

$$\left(2\right)\doteq\left(I_x-\beta C^R\right)^{-1}=I_z+\Lambda^T\Lambda \tag{B.18}$$

and

$$\left(3\right)\doteq\bar{\beta}=\beta\sigma\doteq\sigma\left(C^R\right)^T\left[C^R\left(C^R\right)^T+\sigma^2 I_x\right]^{-1}=\Lambda^T\left[I_x+\Lambda\Lambda^T\right]^{-1};\quad \Lambda\doteq C^R/\sigma \tag{B.19}$$

and

$$\left(1\right)\doteq\left(\bar{\beta}\right)^T=\left\{\Lambda^T\left[I_x+\Lambda\Lambda^T\right]^{-1}\right\}^T=\left\{\left[I_x+\Lambda\Lambda^T\right]^{-1}\right\}^T\Lambda=\left[I_x+\Lambda\Lambda^T\right]^{-1}\Lambda\,. \tag{B.20}$$

The last equality following from $\left(A^{-1}\right)^T=\left(A^T\right)^{-1}=\left(A\right)^{-1}$ for any symmetric matrix.

To flip $\Lambda\Lambda^T$ to $\Lambda^T\Lambda$ now use Woodbury [12] in the direction

$$\left(A+UCV\right)^{-1}=A^{-1}-A^{-1}U\left(C^{-1}+VA^{-1}U\right)^{-1}VA^{-1} \tag{B.21}$$

on Eq.(B.19) for (3) and Eq.(B.20) for (1). Note that $\left(1\right)=\left(3\right)^T$.

Setting: $A=I_x;C=I_z;U=\Lambda;V=\Lambda^T$

$$\left(3\right)=\Lambda^T\left(I_x+\Lambda\Lambda^T\right)^{-1}=\Lambda^T\left[I_x-\Lambda\left(I_z+\Lambda^T\Lambda\right)^{-1}\Lambda^T\right]. \tag{B.22}$$

Therefore

$$\left(1\right)=\left[\left(I_x+\Lambda\Lambda^T\right)^{-1}\right]^T\Lambda=\left[\left(I_x+\Lambda\Lambda^T\right)\right]^{-1}\Lambda=\left[I_x-\Lambda\left(I_z+\Lambda^T\Lambda\right)^{-1}\Lambda^T\right]\Lambda\,. \tag{B.23}$$

Thus, along with Eq. (B.7) for $\left(2\right)$ in Eq. (B.17)

$$\left(\beta x-\mu_z^V(x)\right)^T\left(I_x-\beta C^R\right)^{-1}\left(\beta x-\mu_z^V(x)\right)=\left(\overline{x}-\overline{\mu}_z^V(x)\right)^T S\left(\overline{x}-\overline{\mu}_z^V(x)\right) \quad \text{(B.24)}$$

with

$$S \doteq\left[I_x-\Lambda\left(I_z+\Lambda^T\Lambda\right)^{-1}\Lambda^T\right]\left[\Lambda\left(I_z+\Lambda^T\Lambda\right)\Lambda^T\right]\left[I_x-\Lambda\left(I_z+\Lambda^T\Lambda\right)^{-1}\Lambda^T\right]. \quad \text{(B.25)}$$

$S$ only requires inversion of $\left(I_z+\Lambda^T\Lambda\right)$ that is readily evaluated in the latent space. Use of the SVD on $\Lambda$ eliminates even this requirement with the properties in Eq. (B.8).

Recognizing that $\left(I_z+L^2\right)^{\pm 1}$ are diagonal matrices, and that $V=\left(V^{-1}\right)^{-1}=\left(V^T\right)^{-1}$

$$\begin{aligned}&\left(I_z+\Lambda^T\Lambda\right)^{+1}=\left(I_z+VL^2V^T\right)^{+1}=V\left(I_z+L^2\right)^{+1}V^T\\&\text{and}\\&\left(I_z+\Lambda^T\Lambda\right)^{-1}=\left(I_z+VL^2V^T\right)^{-1}=\left[V\left(I_z+L^2\right)V^T\right]^{-1}\\&=\left(V^T\right)^{-1}\left(I_z+L^2\right)^{-1}V^{-1}=V\left(I_z+L^2\right)^{-1}V^T .\end{aligned} \quad \text{(B.26)}$$

Thus

$$\begin{aligned}&\Lambda\left[I_z+\Lambda^T\Lambda\right]^{\pm 1}\Lambda^T=ULV^TV\left[I_z+L^2\right]^{\pm 1}V^TVLU^T\\&=UL\left[I_z+L^2\right]^{\pm 1}LU^T=UG_d^{\pm}U^T;\quad G_d^{\pm}\doteq L\left[I_z+L^2\right]^{\pm 1}L.\end{aligned} \quad \text{(B.27)}$$

Eq. (B.27) defines diagonal matrices $G_d^{\pm 1}$ with elements

$$\left(G_d^{\pm}\right)_{kl}\doteq\delta_{kl}\lambda_l^2\left(1+\lambda_l^2\right)^{\pm 1}. \quad \text{(B.28)}$$

Eqs. (B.25) and (B.27) permits expression of $S$ as

$$\begin{aligned}S&=\left[\left(I_x-UG_d^-U^T\right)U\right]G_d^+\left[U^T\left(I_x-UG_d^-U^T\right)\right]=\left[U-UG_d^-U^TU\right]G_d^+\left[U^T-U^TUG_d^-U^T\right]\\&=\left[U-UG_d^-I_z\right]G_d^+\left[U^T-I_zG_d^-U^T\right]=U\left[I_z-G_d^-\right]G_d^+\left[I_z-G_d^-\right]U^T.\end{aligned} \quad \text{(B.29)}$$

The matrix $I_z-G_d^-$ has elements

$$\left(I_z-G_d^-\right)_{kl}=\delta_{kl}-\delta_{kl}\frac{\lambda_l^2}{1+\lambda_l^2}=\delta_{kl}\frac{1}{1+\lambda_l^2}. \quad \text{(B.30)}$$

Define a diagonal matrix $H_d$ with elements

$$\left(H_d\right)_{kl}\doteq\left(\left[I_z-G_d^-\right]G_d^+\left[I_z-G_d^-\right]\right)_{kl}=\delta_{kl}\frac{1}{1+\lambda_l^2}\lambda_l^2\left(1+\lambda_l^2\right)\frac{1}{1+\lambda_l^2}=\delta_{kl}\frac{\lambda_l^2}{1+\lambda_l^2} \quad \text{(B.31)}$$

Thus,

$$\left(\beta x-\mu_z^V(x)\right)^T\left(I_x-\beta C^R\right)^{-1}\left(\beta x-\mu_z^V(x)\right)=\left(\overline{x}-\overline{\mu}_z^V(x)\right)^T UH_dU^T\left(\overline{x}-\overline{\mu}_z^V(x)\right) \tag{B.32}$$

with the elements of $H_d$ given in Eq (B.31). In more explicit fashion, introduce the $N_z$ eigenvectors $u_l\left(N_x\times 1\right)$ of $U\left(N_x\times N_z\right)$

$$U\doteq\left(u_1,u_2,...,u_{N_z}\right)\quad l=1,2,..,N_z\,. \tag{B.33}$$

Then, the RHS of (B.32) reduces to

$$\left(\overline{x}-\overline{\mu}_z^V(x)\right)^T UH_dU^T\left(\overline{x}-\overline{\mu}_z^V(x)\right)=\sum_{l=1}^{N_z}\left[\left(\overline{x}-\overline{\mu}_z^V(x)\right)^T\bullet u_l\right]\left(H_d\right)_{ll}\left[u_l^{\,T}\bullet\left(\overline{x}-\overline{\mu}_z^V(x)\right)\right] \tag{B.34}$$

with $\bullet$ indicating scalar product of vectors.

Finally, also needed is an explicit expression for $\overline{\mu}_z^V(x)\doteq\overline{\beta}^{-1}\bullet\mu_z^V(x)/\sigma$ that requires $\overline{\beta}^{-1}$ that, from Eq. (B.19), is

$$\overline{\beta}^{-1}=\left(\Lambda^T\left[I_x+\Lambda\Lambda^T\right]^{-1}\right)^{-1}=\left[I_x+\Lambda\Lambda^T\right]\left(\Lambda^T\right)^{-1}. \tag{B.35}$$

The $\left[I_x+\Lambda\Lambda^T\right]$ term

$$\left[I_x+\Lambda\Lambda^T\right]=I_x+ULV^T\left(ULV^T\right)^T=I_x+ULV^TVLU^T=I_x+ULLU^T=I_x+UL\left(UL\right)^T \tag{B.36}$$

just involves SVD multiplies. The inverse of $\Lambda^T$ is

$$\left(\Lambda^T\right)^{-1}=\left(VLU^T\right)^{-1}=\left(U^T\right)^{-1}\left(L\right)^{-1}\left(V\right)^{-1}=UL^{-1}V^T, \tag{B.37}$$

using column orthogonality of $U$ and $V$. Thus

$$\overline{\beta}^{-1}=\left[I_x+\Lambda\Lambda^T\right]\left(\Lambda^T\right)^{-1}=\left[I_x+UL\left(UL\right)^T\right]UL^{-1}V^T \tag{B.38}$$

and

$$\overline{\mu}_z^V(x)\doteq\overline{\beta}^{-1}\bullet\mu_z^V(x)/\sigma=\left\{\left[I_x+UL\left(UL\right)^T\right]UL^{-1}V^T\right\}\bullet\mu_z^V(x)/\sigma. \tag{B.39}$$

The transpose is

$$\begin{aligned}\left(\overline{\mu}_z^V(x)\right)^T&=\left(\overline{\beta}^{-1}\,\mu_z^V(x)/\sigma\right)^T=\left(\mu_z^V(x)/\sigma\right)^T\left(\overline{\beta}^{-1}\right)^T\\&=\left(\mu_z^V(x)/\sigma\right)^T\left\{\left[I_x+UL\left(UL\right)^T\right]UL^{-1}V^T\right\}^T\\&=\left(\mu_z^V(x)/\sigma\right)^T\left\{VL^{-1}U^T\left[I_x+UL\left(UL\right)^T\right]\right\}.\end{aligned} \tag{B.40}$$

In summary, putting together Eq. (B.14) for the log trace terms and Eqs. (B.31-34) for the scalar term, by using the SVD of the Roweis covariance matrix, provides $D\left[V_{\varphi_V}\left(z\mid x\right)\|W\left(z\mid x\right)\right]$. The means and variances of $V_{\varphi_V}\left(z\mid x\right)$ are to be learned.

**Appendix C**

We show that $Y-N_z\geq 0$ where

$$Y\doteq\log\left(\left|\Sigma_J\right|\left|\Sigma_I\right|^{-1}\right)+\mathrm{Tr}\left(\Sigma_J^{-1}\Sigma_I\right)\tag{C.1}$$

for arbitrary covariance matrices, $\Sigma_I$ and $\Sigma_J$.

Write $Y$ as

$$Y=\mathrm{Tr}\left(\Sigma_J^{-1}\Sigma_I\right)-\log\left(\left|\Sigma_J\right|^{-1}\left|\Sigma_I\right|\right)=\mathrm{Tr}\left(\Sigma_J^{-1}\Sigma_I\right)-\log\left(\left|\Sigma_J^{-1}\right|\left|\Sigma_I\right|\right)=\mathrm{Tr}\left(\Sigma_J^{-1}\Sigma_I\right)-\log\left(\left|\Sigma_J^{-1}\Sigma_I\right|\right)\tag{C.2}$$

where for second equality we used $\left|\Sigma_J\right|^{-1}=\left|\Sigma_J{}^{-1}\right|$ and, for the third, that $\left|\Sigma_J^{-1}\right|\left|\Sigma_I\right|=\left|\Sigma_J^{-1}\Sigma_I\right|$.

That lets us define one composite matrix $\Sigma_{I/J}\doteq\Sigma_J^{-1}\Sigma_I$ and write

$$Y=\mathrm{Tr}\left(\Sigma_{I/J}\right)-\log\left(\left|\Sigma_{I/J}\right|\right);\quad \Sigma_{I/J}\doteq\Sigma_J^{-1}\Sigma_I\,.\tag{C.3}$$

A useful identity that follows from Jacobi's formula [13] for an invertible matrix such as $\Sigma_{I/J}$ is

$$\mathrm{Tr}\Sigma_{I/J}=\log e^{\mathrm{Tr}\Sigma_{I/J}}=\log\left|e^{\Sigma_{I/J}}\right|.\tag{C.4}$$

Thus, from Eqs. (C.4) and (C.3),

$$Y=\log\left|e^{\Sigma_{I/J}}\right|-\log\left|\Sigma_{I/J}\right|=\log\left[\left|e^{\Sigma_{I/J}}\right|\Big/\left|\Sigma_{I/J}\right|\right].\tag{C.5}$$

Matrices $\Sigma_I$ and $\Sigma_J$ that contribute to $\Sigma_{I/J}$ are symmetric and therefore positive definite from their definitions as covariance matrices, i.e. $\Sigma_x=\mathbb{E}\left[\left(x-\mu_x\right)\left(x-\mu_x\right)^T\right]$. As such, their inverses and powers are well-defined. However, in general, the product of symmetric positive definite matrices is neither symmetric nor positive definite unless the product matrix is similar to a symmetric matrix. If there exists an (invertible) matrix $P$ such that $D=P^{-1}CP$ and $D$ is symmetric then $C$ and $D$ are similar matrices [14] and consequently 1) they have the same eigenvalues and 2) are positive definite matrices with postive eigenvalues.

To show that $\Sigma_{I/J}$ is similar to a symmetric matrix set $C=\Sigma_{I/J}=\Sigma_J^{-1}\Sigma_I$ and $P=\Sigma_I^{-1/2}$. Then

$$D = \Sigma_I^{1/2}\left(\Sigma_J^{-1}\Sigma_I\right)\Sigma_I^{-1/2} = \Sigma_I^{1/2}\Sigma_J^{-1}\Sigma_I^{1/2} \tag{C.6}$$

and $D$ is a symmetric matrix:

$$D^T = \left(\Sigma_I^{1/2}\Sigma_J^{-1}\Sigma_I^{1/2}\right)^T = \left(\Sigma_I^{1/2}\right)^T\left(\Sigma_J^{-1}\right)^T\left(\Sigma_I^{1/2}\right)^T = \left(\Sigma_I^{1/2}\Sigma_J^{-1}\Sigma_I^{1/2}\right) = D\,. \tag{C.7}$$

Thus matrix $C = \Sigma_{I/J}$ is similar to the symmetric matrix $D \doteq \Sigma_I^{1/2}\Sigma_J^{-1}\Sigma_I^{1/2}$.

That the desired eigenvalues of $C$ have the same eigenvalues as $D$ follows from [14]:

$$\begin{aligned}
&D = P^{-1}CP \Leftrightarrow PDP^{-1} = C \\
&\text{If } C\upsilon_l = \lambda_l \upsilon_l \;\text{ then }\; PDP^{-1}\upsilon_l = \lambda_l \upsilon_l \Leftrightarrow DP^{-1}\upsilon_l = \lambda_l P^{-1}\upsilon_l \\
&\text{or }\; D\left(P^{-1}\upsilon_l\right) = \lambda_l\left(P^{-1}\upsilon_l\right).
\end{aligned} \tag{C.8}$$

So, if $\upsilon_l$ is an eigenvector of $C = \Sigma_{I/J}$ with eigenvalue $\lambda_l$ then $P^{-1}\upsilon_l$ is an eigenvector of $D = \Sigma_I^{1/2}\Sigma_J^{-1}\Sigma_I^{1/2}$ with eigenvalue $\lambda_l$. $D$ is positive definite and its eigenvalues $\lambda_l^{\Sigma_{I/J}}$, that are the same as those of $\Sigma_{I/J}$, are positive, providing a positive determinant:

$$\left|\Sigma_{I/J}\right| = \prod_{l=1}^{N_z} \lambda_l^{\Sigma_{I/J}} > 0\,. \tag{C.9}$$

Now note that from Eq (C.4), $\left|e^{\Sigma_{I/J}}\right| = e^{\mathrm{Tr}\Sigma_{I/J}}$

and

$$\mathrm{Tr}D = \mathrm{Tr}\left(\Sigma_I^{1/2}\Sigma_J^{-1}\Sigma_I^{1/2}\right) = \mathrm{Tr}\left(\Sigma_I\Sigma_J^{-1}\right) = \mathrm{Tr}\Sigma_{I/J} = \sum_{l=1}^{N_z} \lambda_l^{\Sigma_{I/J}}\,. \tag{C.10}$$

Therefore

$$\left|e^{\Sigma_{I/J}}\right| = \exp\sum_{l=1}^{N_z} \lambda_l^{\Sigma_{I/J}}; \quad \log\left|e^{\Sigma_{I/J}}\right| = \sum_{l=1}^{N_z} \lambda_l^{\Sigma_{I/J}}\,. \tag{C.11}$$

Using these results to express the determinants in $Y$ in terms of the (positive) eigenvalues $\lambda_l^{\Sigma_{I/J}}$ of $\Sigma_{I/J}$ we get the non-negative result:

$$Y - N_z = \sum_{l=1}^{N_z}\left(\lambda_l^{\Sigma_{I/J}} - 1 - \log\lambda_l^{\Sigma_{I/J}}\right) \geq 0\,. \tag{C.12}$$

This result follows as both $\Sigma_I$ and $\Sigma_J$ are covariances that are represented by symmetric positive definite matrices and that $\Sigma_{I/J} = \Sigma_J^{-1}\Sigma_I$ is similar to the symmetric matrix $\Sigma_I^{1/2}\Sigma_J^{-1}\Sigma_I^{1/2}$ that has positive eigenvalues.